# Vector Symbolic Architectures Answer Jackendoff's Challenges for Cognitive Neuroscience

Ross W. Gayler (r.gayler@mbox.com.au)

**Abstract**

Jackendoff (2002) posed four challenges that linguistic combinatoriality and rules of language present to theories of brain function. The essence of these problems is the question of how to neurally instantiate the rapid construction and transformation of the compositional structures that are typically taken to be the domain of symbolic processing. He contended that typical connectionist approaches fail to meet these challenges and that the dialogue between linguistic theory and cognitive neuroscience will be relatively unproductive until the importance of these problems is widely recognised and the challenges answered by some technical innovation in connectionist modelling. This paper claims that a little-known family of connectionist models (Vector Symbolic Architectures) are able to meet Jackendoff's challenges.

## Introduction

Jackendoff (2002) has posed four linguistic challenges for cognitive neuroscience. He holds that language is a mental phenomenon and that linguistic functionality must be neurally instantiated. However, "although a great deal is known about functional localization of various aspects of language in the brain … nothing at all is known about how neurons instantiate the details of rules of grammar " (Jackendoff, 2002, p. 58).

This lack of progress may be due to the cognitive neuroscientists' choice of tools. "The term *connectionism* has become synonymous with a single kind of network model [(the multilayer perceptron)] … that uses a learning algorithm known as *back-propagation*" (Marcus, 2001, p. xii). In formulating his challenges, Jackendoff draws heavily on Marcus, who argues that multilayer perceptrons are incapable of symbol manipulation. Marcus stresses that he is not an anti-connectionist and "suggest[s] that adequate models of cognition most likely lie in a different, less explored part of the space of possible [connectionist] models"

The lack of progress by cognitive neuroscientists may also be due to attempting to solve the wrong problems, because of holding naive views of linguistic phenomena (Jackendoff, 2002, chap. 3, notes 17 & 21). This apparent naivete may be a conscious research strategy, to start simple, with the intent that current solutions will scale up to ultimate needs. However, this strategy must fail if the initially chosen problems are not actually core to the ultimate problems.

Jackendoff's challenges are an attempt to counter this risk by focussing attention on functionality that he sees as central to all linguistic phenomena. The need for these challenges is shown by the lack of connectionist models able to deal with them successfully and the fact that, despite this lack of success, the challenges "have not been widely recognized [as such] in the cognitive neuroscience community" (Jackendoff, 2002, p. 58). Feldman (2002) broadcast these challenges to the neural network modelling community via the Connectionists Mailing List. The few responses he received were unable to convince Feldman that any standard connectionist techniques would meet Jackendoff's challenges.

The challenges are not exclusively linguistic and are arguably fundamental to all cognition. They are "not something that can simply be disregarded by ignoring language, … but they certainly come to the fore in dealing with the linguistic phenomena that linguists deal with every day" (Jackendoff, 2002, p. 67).

Vector Symbolic Architectures (VSAs) are a little-known class of connectionist models that can directly implement functions usually taken to form the kernel of symbolic processing (Gayler, 1998; Kanerva, 1997; Plate, 1994; Rachkovskij & Kussul 2001). They are an enhancement of tensor product variable binding networks (Smolensky, 1990). Like tensor product networks, VSAs can create and manipulate recursively-structured representations in a natural and direct connectionist fashion without requiring training. However, unlike tensor product networks, VSAs afford a practical basis for implementations because they require only fixed dimension vector representations. The fact that VSAs relate directly, without training, to both simple, practical, vector implementations and core symbolic processing functionality suggests that they would provide a fruitful connectionist basis for the implementation of cognitive functionality.

The approach taken in this paper is constrained by the required brevity. Jackendoff's analysis of the linguistic challenges and the inadequacy of typical cognitive neuroscience models is accepted at face value. The range and properties of VSAs are discussed briefly. The presentation of VSAs and their comparison with other connectionist approaches is minimal, as both topics are adequately documented elsewhere. Then I deal with each of Jackendoff's challenges in turn, showing how they are answered by VSAs.

## Vector Symbolic Architectures

The name "Vector Symbolic Architectures" has been invented to cover a family of related approaches. These approaches are easily implemented as connectionist systems and share a commitment to algebraic operations on distributed representations over high-dimensional vectors. Conceptually, all these approaches are descended from Smolensky's (1990) tensor product variable binding networks. Smolensky demonstrated that variable/value binding and the representation and manipulation of complex nested structures using connectionist methods is possible.

His tensor product approach relies on algebraic operations with simple connectionist implementations. This results in structural manipulation occurring in a single pass through the system, thus avoiding the need for prolonged learning (unlike multilayer perceptrons with back-propagation). Unfortunately, tensor product implementations are thoroughly impractical because the vector dimensionality increases exponentially with the depth of the structures to be represented. Consequently, tensor product binding has been little used in subsequent research.

VSAs (Gayler, 1998; Kanerva, 1997; Plate, 1994; Rachkovskij & Kussul, 2001) retain the advantages of tensor product binding while avoiding the problem of increasing vector dimensionality. In all connectionist systems, entities are represented by vectors. (The activity levels of a set of connectionist units are construed mathematically as a vector.) In tensor product binding, the representation of the association (or binding) of two entities is created as the outer product of the vectors representing the two entities. Thus, if the entity vectors are of dimensionality $n$, the outer product will be of dimensionality $n^2$. VSAs overcome this problem of increasing dimensionality by applying a function to the $n^2$ elements of the outer product to yield a resultant vector of dimensionality $n$. Thus all structures, whether atomic or complex, are represented by vectors of the same dimensionality.

The application of VSAs to cognitive problems is arguably more complex than the application of conventional multilayer perceptrons with back-propagation. Three separate levels of description are required. These levels are: the vector representation, the representational architecture and the cognitive architecture. This paper, like other papers on VSAs, deals primarily with the vector representation level and only touches on the representational architecture to the minimum extent necessary. The contention is that VSAs examined at these levels are manifestly better suited to cognitive tasks than the connectionist alternatives and that the extension of the analyses through the representational and cognitive architecture levels can be reasonably expected as a consequence of the ongoing VSA research program.

The currently available VSAs employ three types of operation on vectors: a multiplication-like operator, an addition-like operator, and a permutation-like operator. The precise choice for each operator varies by VSA. Gayler (1998) and Plate (1997) compare the implementations of some VSAs. The multiplication-like operation is used to associate or bind vectors. The addition-like operation is used to superpose vectors or add them to a set. The permutation-like operation is used to quote or protect vectors from the other operations.

For the sake of concreteness, in examples I will use the MAP (Multiply, Add, Permute) Coding scheme described in Gayler (1998). In MAP Coding, elementwise multiplication is used to implement the binding of vectors; elementwise addition is used to implement the superposition of vectors; and permutation of the elements is used to implement quotation of vectors. These operations can be used to compose, decompose, and manipulate complex structures without requiring any training (Gayler, 1998). Analogous descriptions of other VSAs can be found in Kanerva (1997), Plate (1994), and Rachkovskij & Kussul (2001).

The description of VSAs has so far been at the vector representation level, being purely in terms of the content of vectors. Corresponding to each operation there is a representational architecture primitive. For example, to implement binding there would be a layer of connectionist units taking two inputs each and computing the product of those inputs. These architectural primitives are combined in some fixed circuit to yield a complete representational architecture. The cognitive architecture level deals with how a cognitive problem is represented in terms of vector representations and prior knowledge in order to obtain the desired behaviour from a given representational architecture.

## Challenge 1: The Massiveness of the Binding Problem

Jackendoff's first challenge arises from the observation that linguistic representations must be composed from component representations. "The need for combining independent bits into a single coherent percept has been recognized in the theory of vision under the name of the *binding problem*" (Jackendoff, 2002, p. 59). The typical presentation of the binding problem in vision is given in terms of associating attributes with an object, for example, representing an object as both red and square. The challenge in this problem becomes more obvious when there is more than one object, for example, a red square and a blue circle. It is clear that a mechanism is needed to ensure that the correct attributes are associated with each object. Nonetheless, the typical presentation of the problem is given in terms of only a few objects and attributes. Jackendoff then shows that the binding problem is much larger in linguistics.

Jackendoff refers repeatedly to the example sentence "The little star's beside a big star" (2002, p. 5). He gives a representation of the structure encoded in the sentence on the phonological, syntactic, semantic/conceptual, and spatial levels (fig. 1.1). My informal count of this structure shows approximately 130 tokens and 160 relations between tokens. The number of bindings involved in the composition of the structure corresponding to this simple sentence is obviously orders of magnitude larger than in the typical visual example of binding.

There are two other aspects of the binding problem that must be mentioned: novelty and speed. Although the individual components at the lowest level will be familiar, the total composite will frequently be novel. Furthermore, this novel composite must be constructed in the time it takes to comprehend a single sentence. Thus, approaches relying on extensive training to construct each composite structure are implausible.

**Response 1**
Binding in MAP Coding is implemented as the elementwise product of the vectors to be bound.
$$bind(a,b) = a*b$$
Unbinding in VSAs is implemented as the binding of the inverse of the cue to the trace.
$$unbind(cue, trace) = bind(inverse(cue), trace)$$
These functions yield the desired behaviour for binding and unbinding in that a cue can be used to retrieve components from a bound composite.
$$unbind(a, bind(a,b)) = b$$
$$unbind(b, bind(a,b)) = a$$
The mathematical details of binding and unbinding in VSAs and the extension to noisy cues and traces can be found in Plate (1994) and Kanerva (1997).

In MAP Coding (Gayler, 1998) and Spatter Coding (Kanerva, 1997) each vector is its own inverse with respect to binding
$$bind(a,a) = a*a = 1$$
and no separate inverse function is required. In this case, the multiplication-like operation can be construed as implementing a number of functions (binding, unbinding, and others), depending on the relationship between the vectors being multiplied (Gayler & Wales, 2000).

Binding is implemented at the representational architecture level as a simple primitive: a layer of connectionist units computing the elementwise product of the inputs. The binding occurs in a single pass through the units as a consequence of the algebraic relationship between the inputs and outputs. Thus, it is fast and oblivious to the novelty or familiarity of the items to be bound. Another consequence of the blindness of binding to the contents of the vectors to be bound is that the entities to be bound may be composite structures in their own right. This allows the construction of complex recursive structures (Plate, 1994; Smolensky, 1990).

The vector representation level of VSAs clearly has the capability to deal, in principle, with the binding problem. The representational architecture primitives have the required characteristics of speed and ability to cope with novelty. It is trivially easy to construct a representational architecture that implements a fixed compositional template. However, there is no currently available representational architecture and cognitive architecture that implements fully variable, input driven composition. My current research (Gayler & Wales, 2000) is aimed at this objective.

Connectionist recurrent associative memories (e.g. Anderson, Silverstein, Ritz, & Jones, 1977) work by creating attractors corresponding to each of the items to be remembered. My research aims to create an enhanced recurrent associative memory that has attractors corresponding to novel, valid compositions of familiar items. Such a memory would be able to recognise novel composites by retrieving the components and generating mappings between them. A similar approach, not based on VSAs, has already been successfully demonstrated in the visual domain (Arathorn, 2002).

## Challenge 2: The Problem of 2

Put simply, this challenge asks how multiple instances of the same token are instantiated. In terms of the example sentence: How are the "little star" and "big star" instantiated so that they are both stars, yet distinguishable?

**Response 2**
In VSAs, superposition (implemented by the addition-like operator) is used to represent multiple items in the same representational space (over the same set of connectionist units). The primitives at the representational architecture level can be used to add items to a superposition and operate on a set of superposed items simultaneously (Kanerva, 1997; Plate, 1994).

In VSAs, identical vector values can only be kept distinct by representing them in separate parts of the representational architecture. That is, in a given representational space (a set of connectionist units) only vector values exist. A superposition of the same vector with itself is merely a rescaling of that vector, not a representation of two distinct entities.
$$star + star = 2\ star$$
In order for multiple instances to be represented simultaneously in a superposition those instances must be distinct. In the external world, the little star is distinct from the big star by virtue of (among other things) the difference in size. This difference can be captured by representing each star with a composite structure that includes the size attribute. If the composite representations are distinct they can be superposed without losing their identities.

At the vector representation level, superposition is a similarity preserving operation, whereas binding is a similarity destroying operation (where the similarity of two vectors is measured by their dot product).

$$(a + b) . a > 0$$
$$(a * b) . a = 0$$

If we were to encode the representation of each star as the binding of a size token with the star token, the representations of the two stars would be dissimilar.

$$little . big = 0$$
$$(little * star) . (big * star) = 0$$

These two vectors (**little*star** and **big*star**) may be safely superposed without losing their identities.

Unfortunately, this naive encoding scheme does not completely solve the problem of representing multiple instances. Consider what happens when we want to represent a little star and a red star using the same encoding scheme:

$$(little * star) + (red * star) = (little + red) * star$$

There is nothing in the encoding to differentiate between two stars, one little and one red, and one little red star.

One way to overcome this problem is to use a more frame-like representation with role:filler pairs. The two stars could be represented abstractly as:

**{frame-1 is-a:star size:little}**
**{frame-2 is-a:star colour:red}**

with corresponding vector representations:

**frame-1 * (is-a*star + size*little)**
**= frame-1*is-a*star + frame-1*size*little**

**frame-2 * (is-a*star + colour*red)**
**= frame-2*is-a*star + frame-2*colour*red**

When these two representations are superposed the properties of the stars are kept separate.

This encoding begs the question of the origin of the frame identity tokens (**frame-1** and **frame-2**). A good solution to this problem is to calculate the frame identity as the permutation of the frame contents (Gayler, 1998).

**make-frame(a) = bind(P(a), a)**

where **P()** is the permutation-like operator. Then the little star is represented as:

**make-frame(is-a*star + size*little)**
**= P(is-a*star + size*little)*(is-a*star + size*little)**
**= P(is-a*star + size*little)*is-a*star**
**+ P(is-a*star + size*little)*size*little**

That is, each role:filler pair is bound with every other role:filler pair in the same frame, thus capturing the holistic nature of the frame as an entity. Note that frames with identical contents would have identical vector representations and thus be indistinguishable, which seems quite reasonable.

Construction of a representational architecture to implement the **make-frame** function is trivial. This architecture consists of only one permutation operator and one binding operator.

## Challenge 3: The Problem of Variables

The third challenge arises from the productivity of language. Language users are able to recognise and generate an infinite variety of utterances using only finite resources. This productivity is construed as arising from the use of variables; placeholders which may contain arbitrary values (Jackendoff, 2002, p. 64).

A rule of a grammar can be construed as a template with variables. When those variables are instantiated, the remainder of the template relates their values to each other. Thus, a rule can be seen as a mechanism for constructing or recognising a composite structure, and the variables of the rule as markers for the components to be productively replaced.

The variables in rules are generally "typed". That is, the set of values with which a variable may be instantiated must be constrained. The type of a variable is the constraint on its possible values. Traditionally, rules are seen as quite distinct from the structures that they operate on and types are annotations on variables in rules. However, this distinction between rules and structures can be erased. In a thoroughly lexicalised grammar, the information in rules is expressed as structural fragments (like the structures operated on by the rules) which are combined by "the only *procedural* rule … UNIFY ('clip' structures together)" (Jackendoff, 2002, p. 182). In such a grammar, the compatibility constraints (types on variables) can be captured in the structure of the fragments.

### Response 3

Smolensky (1990) demonstrated that tensor product connectionist systems can implement variables. VSAs inherit this capability. Instantiation of a variable can be implemented by binding a vector representing the variable with a vector representing the value. The value can be retrieved from the variable by probing the binding with the identity of the variable as a cue.

**unbind(<u>variable</u>, bind(variable, value))**
**= variable * variable * value**
**= <u>value</u>**

Note that there is nothing special about the vector representing the identity of the variable. It is just another vector and could as easily be the representation of a complex structure as the representation of an atomic token. This allows complex structures to be interpreted as overlapping networks of variable/value bindings, for example:

**a*b*c*d = a*(b*c*d) = (a*b)*(c*d) = (a*b*c)*d**

Smolensky has used the traditional, imperative programming language interpretation of a variable as a location to hold a value. However, there is an alternative interpretation of the concept of a variable that is based on declarative programming languages and is more congenial to constraint-based grammar formalisms (Copestake, 2002). This interpretation treats variables as targets for substitution. This interpretation can also be implemented with VSAs.

A substitution may be represented as the binding of the representations to be substituted for each other (Gayler, 2000).

**make-substitution(a, b) = bind(a, b)**

The substitution may be applied to a structure by binding the substitution with the structure.

**apply-substitution(substitution, structure)
= bind(substitution, structure)**

apply-substitution(make-substitution(x,b), <u>bind(x,y)</u>)
= (x*b) * (x*y)
= b*y
= <u>bind(b,y)</u>

Given this capability, any component of a structure can act as a variable (a target for substitution). The only constraint is that if the same component has multiple occurrences in the one structure the substitution is applied identically to all occurrences.

apply-substitution(make-substitution(x,b), <u>(x*y + x*z)</u>)
= <u>(b*y + b*z)</u>

This suggests a style of processing in which only literal episodes are stored and rules arise as statistical regularities of potential substitutions across those episodes, similar to the Data Oriented Processing of Bod and Scha (1997).

## Challenge 4: Binding in Working Memory vs Long-Term Memory

Jackendoff's fourth challenge concerns the transparency of the boundary between working memory and long-term memory. He argues that linguistic tasks require the same structures to be instantiated in working memory and long-term memory and that the two instantiations should be functionally equivalent (2002, p. 65).

Two aspects of typical connectionist implementations suggest a lack of equivalence between working memory and long-term memory representations. Working memory representations are typically implemented using activation levels and (possibly) temporal synchrony, whereas long-term memory is implemented using synaptic connectivity. The disparity of implementation media suggests that it would be difficult to achieve functional equivalence. The other aspect is speed, again. Linguistic phenomena require single trial learning and single trial learning seems incompatible with gradual strengthening of synaptic connectivity.

## Response 4

Jackendoff has cast this problem as arising from the difference in physical implementation of working memory and long-term memory. I recast this problem as arising from differences at the vector representation level. In typical connectionist systems, working memory items are represented as vectors (of activations) and long-term memory items are represented as matrices (of synaptic connectivities).

The crucial difference here is not the physical implementation (activations versus connectivities), but the logical form of the representations (vectors versus matrices). Of course, the synaptic matrices can be reinterpreted as vectors, but they will be of different dimensionality to the activation vectors. That is, the working memory items and long-term memory items exist in different, incommensurable representational spaces. This is hardly surprising, as the synaptic connectivities can be construed as relations between working memory items. They are the mechanism by which items in working memory create other items in working memory.

In VSAs, working memory items and long-term memory items have the same form at the vector representation level. They are both represented by vectors of the same dimensionality. Those vectors can represent items and relationships between items both in working memory and long-term memory. This equivalence of logical form makes the boundary between working memory and long-term memory transparent.

The differences in physical implementation are reflected not in what can be represented but in the persistence of those representations and their ability to interact with each other. Working memory items are able to interact with each other and items in long-term memory, whereas long-term memory items are only able to interact with each other via items in working memory. This issue has been discussed with respect to a different set of challenges in Gayler and Wales (1998).

The issue of speed of learning has two components: speed of binding and speed of laying down a trace. As discussed in response to the first challenge, binding in VSAs arises from the algebraic properties of the operations and occurs in a single pass through an architectural primitive. Thus, associations can be formed in a single trial.

Separate from that is the issue of whether those associations can be made persistent in a single trial. For items being made persistent in working memory this consists of superposing the new association on the current activity vector. This can be done in a single pass through the architectural primitive that implements the addition-like operation. This could have multiple physical implementations, for example, a direct increment to the activity levels, or short-term potentiation of the representational units. Long-term persistence is achieved in functionally equivalent ways with different implementations, for example, long-term potentiation of representational units, or increments to the synaptic connectivities.

## Conclusion

Jackendoff posed four challenges that are generally relevant to cognition but particularly relevant to language. These challenges identify functional capabilities that are required for language, but arguably not provided by typical connectionist models. VSAs, a little-known family of connectionist architectures, do meet these challenges. Their ability to meet these challenges arises from the algebraic properties of their vector representations. These properties and the fact that they have straightforward physical implementations suggest that VSAs are ideal candidates for cognitive modelling. However, it has to be asked why VSAs are so relatively little-known.

There are multiple reasons that could be presented, but I will offer only one. While VSAs are very easy to work with at the vector representation level and the level of primitives of the representational architecture, they are very difficult to work with at the level of complete representational architectures and cognitive architectures. Typical connectionist architectures rely on training procedures to achieve their effectiveness. However, VSAs provide no opportunity for training to substitute for architectural effectiveness. That is, good performance depends on good design rather than automated training, and this is a harder research task.